# Empowering Smaller Models: Tuning LLaMA and Gemma with Chain-of-Thought for Ukrainian Exam Tasks

**MYKYTA SYROMIATNIKOV[1], VICTORIA RUVINSKAYA[1], NATALIIA KOMLEVA[1]**

[1]Odesa Polytechnic National University. Odesa 65044 Ukraine

Corresponding author: Mykyta Syromiatnikov (e-mail: nik.syromyatnikov@gmail.com).

**ABSTRACT** Leading large language models have demonstrated impressive capabilities in reasoning-intensive tasks, such as standardized educational testing. However, they often require extensive training in low-resource settings with inaccessible infrastructure. Small or compact models, though more efficient, frequently lack sufficient support for underrepresented languages, leaving a performance gap in critical domains. This work explores the potential of parameter-efficient fine-tuning of compact open-weight language models to handle reasoning-intensive tasks in the underrepresented Ukrainian language, building on the findings of the ZNO-Eval benchmark. Parameter-efficient fine-tuning of LLaMA 3.1 (8 billion parameters), LLaMA 3.2 (3 billion parameters), and Gemma 2 (9 billion parameters) models on chain-of-thought solutions resulted in a modest test score improvement of up to 17.4% on complex matching tasks and 1.6% overall compared to tuning on answer letters alone, offering enhanced interpretability and robustness. In addition, the proposed tuning method with joint task topic and step-by-step solution generation outperforms standard chain-of-thought tuning in matching tasks and provides a 5.4% gain over the best LLaMA 3.2 model due to guiding the model to recall and apply domain-relevant information. Contrasting obtained results with zero-shot evaluations of leading open-weight and proprietary models such as Qwen, DeepSeek R1, OpenAI o1 and o3, Gemini, and Claude, highlight that fine-tuning LLaMA and Gemma models with 2,032 step-by-step solutions and 20 to 50 million trainable parameters on a single A100 GPU lets them outperform GPT-4o mini, Mistral Large, and larger open-weight models. This research also evaluates how merging the quantized adapter with the base model influences the generation quality.

## I. INTRODUCTION

In recent years, Large Language Models (LLMs) have demonstrated remarkable proficiency in language understanding tasks, surpassing human-level performance on multiple benchmarks with narrow text understanding tasks, including traditional GLUE [1] and SQuAD [2], as well as challenging reasoning benchmarks like the Massive Multitask Language Understanding (MMLU) benchmark, which encompasses a wide array of subjects and requires advanced reasoning skills [3]. Moreover, in addition to unprecedented accuracy in complex linguistic challenges, giant models like PaLM empower new discoveries in mathematical sciences [4]. At the same time, multimodal solutions like GPT-4o or LLaMA excel in integrating visual and textual information, enabling sophisticated image captioning and visual question-answering [5].

With the advent of the Transformer architecture [6], enhanced attention mechanisms [7], and neural scaling laws [8], language models have revolutionized a wide range of fields, including search, recommendation systems, real-time coding assistance, and even robotics [9], fundamentally reshaping how information is processed and utilized across domains and industries.

Technologies like these serve as a foundation for developing information systems that can be implemented across various domains and integrated with other neural network architectures and machine learning methods. This approach enables the solution of complex, semi-formalized practical tasks that require flexible adaptation and the combination of different intelligent methods [10].

Despite their impressive achievements, LLMs often demonstrate limited capabilities in underrepresented languages. Studies highlight that these models, predominantly trained on high-resource languages like English, struggle to generalize effectively to low-resource languages, resulting in degraded accuracy and robustness [11]. In particular, the ZNO-Eval benchmark based on standardized exams for Ukrainian school graduates showcased zero-shot strength in factual recall and general knowledge across all models in the National multi-subject test (NMT) subsets for history and geography [12]. However, the answer quality drops substantially when dealing with intricate language and specialized terminology of Ukrainian language exam tasks (Table 1). Moreover, the substantial computational resources required to train and deploy large language models introduce additional complexity. The enormous model sizes, exceeding tens or hundreds of billions of parameters, require extensive hardware capabilities, which makes them less suitable for organizations and individuals with limited resources. This scalability issue highlights the importance of more efficient, compact models that deliver relatively high performance without the associated computational overhead.

**Table 1. Sample tasks from Ukrainian language exam along with their English translations**

| Task description | English translation of task description |
|---|---|
| Завдання з вибором однієї правильної відповіді:<br>Суфікс **-ин-** має однакове значення в усіх словах, ОКРІМ<br>А - соломина<br>Б - бадилина<br>В - височина<br>Г - стеблина | Single correct answer task:<br>The suffix **-ин-** has the same meaning in all of the following words EXCEPT<br>A – соломина (straw piece)<br>B – бадилина (leafy-stalk piece)<br>C – височина (height / highland)<br>D – стеблина (stem piece) |
| Завдання на встановлення відповідності (логічні пари):<br>З'ясуйте, якими частинами мови є виділені слова в реченні (цифра позначає наступне слово).<br>Сучасна людина, щоб бути (1)**успішною**, має вчитися (2)**впродовж** (3)**усього** життя, (4)**опановуючи** нові галузі знань.<br>А - займенник<br>Б - прикметник<br>В - форма дієслова (дієприкметник)<br>Г - форма дієслова (дієприслівник)<br>Д - прийменник | Matching task (logical pairs):<br>Determine which parts of speech the highlighted words are in the following sentence (the number indicates the word that follows).<br>A modern person, in order to be (1)**successful**, must keep studying (2)**throughou**t (3)**their** entire life, (4)**mastering** new fields of knowledge.<br>A – pronoun<br>B – adjective<br>C – verb form (participle)<br>D – verb form (adverbial participle)<br>E – preposition |

Due to these challenges, the research community has shifted its focus toward developing smaller language models that maintain competitive performance levels [13]. This ongoing effort includes advancements in model training, such as promising parameter-efficient fine-tuning methods (PEFT) that significantly reduce the number of trainable parameters [14], alongside innovative prompting techniques that augment input with instructions to boost performance [15].

By employing these strategies, smaller models can be fine-tuned to approach or even match the performance of their larger competitors in English tasks while mitigating the computational demands [16]. Consequently, there is a growing interest in exploring whether these efficient models can be improved in low-resource setups for underrepresented languages to achieve comparable performance on specific tasks, thereby democratizing access to AI capabilities.

This paper explores the efficacy of fine-tuning compact open-source language models, specifically LLaMA and Gemma, combining PEFT and prompt tuning methods to enhance performance on Ukrainian exam tasks.

## II. RELATED WORKS
### A. COMPACT LANGUAGE MODELS

Compact language models have gained attention due to their ability to deliver robust performance while requiring fewer computational resources than larger models. With advancements in mobile computing, these compact yet powerful models are increasingly favored for edge device deployment [17]. They offer enhanced privacy and reduced network dependency, making them an attractive option for a wide range of applications. Notable among these compact LLMs are the Gemma 2 and LLaMA 3 model families.

Google's Gemma 2 open-source models are decoder-only large language models designed for text-to-text generation tasks. They are available in multiple parameter sizes, specifically 2 billion (2B), 9 billion (9B), and 27 billion (27B) parameters. The architecture introduces several technical modifications to the Transformer framework, such as interleaving local-global and group-query attention, contributing to improved performance and efficiency [18].

Gemma 2 models have demonstrated exceptional benchmark results across various natural language processing tasks. Notably, these models outperform some larger open models, showcasing their efficiency and effectiveness despite a relatively smaller parameter count. The instruction-tuned variants of Gemma 2 are reliable at following user prompts and generating coherent, contextually relevant responses [18].

The LLaMA series of open-source LLMs, developed by Meta, has seen significant advancements with the introduction of LLaMA 3, LLaMA 3.1, LLaMA 3.2, and LLaMA 3.3 models. These iterations have progressively enhanced capabilities, model sizes, and functionalities to support diverse AI applications.

Released in April 2024, LLaMA 3 marked a substantial upgrade in Meta's language model offerings. It was introduced in two parameter sizes: 8 billion (8B) and 70 billion (70B). The 70B model was trained on approximately 15 trillion tokens, enabling it to outperform competitors like Gemini Pro 1.5 and Claude 3 Sonnet on various benchmarks [19].

In July 2024, Meta released LLaMA 3.1, expanding the model sizes to include 8B, 70B, and a new 405 billion (405B) parameter model. The 405B model featured an extended context window of up to 128,000 tokens, allowing it to process longer inputs effectively. LLaMA 3.1 aimed to boost efficiency, addressing the limitations of its predecessor [20].

The introduction of LLaMA 3.2 in September 2024 brought significant advancements, particularly in multimodal processing. This version included models with 1B, 3B, 11B, and 90B parameters suitable for various use cases. The 11B and 90B parameter models were designed for joint text and image tasks, while the 1B and 3B models were optimized for deployment on edge devices, supporting real-time processing [21].

### B. EFFICIENT FINE-TUNING OF LLM

Fine-tuning large language models usually demands substantial computational resources. Parameter-efficient fine-tuning techniques like Low-Rank Adaptation (LoRA) have been developed to address this. LoRA reduces the number of trainable parameters by introducing trainable low-rank matrices into each layer of the Transformer architecture, allowing for efficient adaptation of pre-trained models to specific tasks without full model retraining [22].

The quantization method is another option to enhance the efficiency of model training or inference by reducing the precision of its weights, thereby decreasing memory usage and increasing computation speed. For example, QLoRA enables fine-tuning of a 65-billion parameter model on a single 48GB GPU while preserving full 16-bit fine-tuning task performance [23]. These advancements make deploying sophisticated LLMs in environments with limited computational resources feasible.

### C. PROMPTING TECHNIQUES

Prompting techniques have become crucial tools for effectively guiding large language models to perform a wide range of natural language processing tasks and produce the desired output. These methods enable users to configure LLMs for specific behavior without modifying their internal parameters, making them suitable for various applications in low-resource environments. A list of common strategies includes the following.

1. Zero-Shot Prompting, where the model is given a task description without any examples and is expected to generate the correct output based solely on the prompt [15], leverages the model's pre-existing knowledge to handle tasks on which it has not explicitly been trained.

2. Few-Shot Prompting involves providing the model with a few input-output examples within the prompt to illustrate the task, enabling it to infer and apply the desired pattern to new inputs and typically improving robustness and accuracy [24].

3. Chain-of-Thought (CoT) Prompting encourages the model to decompose complex problems into a series of intermediate reasoning steps before printing the final answer. This method enhances the model's ability to perform tasks that require logical reasoning and multi-step problem-solving [25].

4. Instruction Prompting empowers the model with explicit instructions or guidelines on how to approach a task. Clear and detailed instructions can significantly improve the model's performance by aligning its outputs with user expectations.

5. Generated Knowledge Prompting involves prompting the model to generate relevant background knowledge before moving to the main task. The model can produce more robust and contextually appropriate responses by first generating relevant information [26].

In general, prompt engineering is a crucial skill for deploying large language models and can effectively guide LLMs toward improved generalization and reduced hallucinations, particularly for underrepresented languages and complex problem domains.

### D. EXPLAINABLE AI

Explainable Artificial Intelligence is a set of information technologies, models, and methods that help users understand and trust the results produced by machine learning algorithms. Some simple models, such as regression or decision trees, can be explained without additional effort. In earlier knowledge-processing approaches, explanations were provided based on the fragments of knowledge used to obtain prediction results [27]. Modern machine learning methods, such as deep neural networks, are often viewed as "black boxes" due to sophisticated inner workings that are hard to interpret. However, even for these complex models, there are now model-agnostic methods and frameworks for explainability. These typically involve three stages of explanation: pre-modeling (which includes dataset explorations of all kinds), during modeling (where explanations become part of the model's internal functioning), and post-modeling (providing explanations for the prediction results) [28].

Attention visualization is another valuable tool for Transformer-based architectures, especially in natural language processing tasks, where it can highlight the input segments with a high effect on the model's outputs. However, an even more promising strategy is the chain-of-thought prompting. CoT not only improves the accuracy of predictions but also explicitly presents intermediate steps to unveil the intuition behind any intermediate decision. This detailed explanation simplifies a deeper evaluation of the model's performance, allowing users to verify that the reasoning aligns with domain-specific rules and principles. This interpretability is crucial for various applications – from academic assessments to healthcare diagnostics – where understanding the motivation behind a decision is no less important than the decision itself.

### E. SOLVING EXAM TASKS WITH LLM

The application of LLMs to standardized exam tasks serves as a vital benchmark for their reasoning abilities. For the English language, benchmarks like MMLU, GSM8K, and BIG-Bench provide comprehensive datasets for evaluating model performance on academic examinations:

- MMLU (Massive Multitask Language Understanding) benchmark assesses a model's knowledge and reasoning abilities across over 57 tasks spanning diverse academic disciplines, including mathematics, history, and literature [3];

- GSM8K (Grade School Math 8K) is a widely used benchmark for evaluating multi-step reasoning and arithmetic capabilities, consisting of 8,000 math problems designed to test logical deduction and numerical accuracy [29];

- BIG-Bench (Beyond the Imitation Game Benchmark) – a large-scale benchmark featuring over 200 diverse tasks, such as logic, mathematics, common sense reasoning, and language generation, aimed at pushing models to exhibit deeper cognitive understanding and reasoning [30].

For the Ukrainian language, the ZNO-Eval benchmark with real exam tasks from Ukraine's standardized educational testing system, including the External Independent Evaluation and the National Multi-subject Test, comprises single-answer options, matching, correct sequence, and open-ended questions across diverse subjects, delivering a thorough analysis of proprietary LLMs' reasoning capabilities in Ukrainian [12].

At the same time, the UNLP 2024 Shared Task initiative made significant contributions to the benchmarking of open-weight models [31]. This initiative aimed to support the development of models with a deep understanding of the Ukrainian language, literature, and history. It showcased fine-tuning results for numerous promising models and strategies, highlighting advancements in adapting LLMs for Ukrainian-specific tasks [32].

The ZNO-Vision benchmark further extends the evaluation of large language models to multimodal contexts by incorporating over 4,300 expert-crafted questions spanning 12 academic disciplines, including mathematics, physics, chemistry, and humanities [33]. This dataset includes visual elements, enabling the assessment of models' capabilities in handling both text and images.

However, both the Shared Task and ZNO-Vision evaluations focused solely on questions with a single correct answer. In contrast, ZNO-Eval tasks involving matching or correct sequences could provide valuable insights into tuning strategies, as they offer a deeper assessment of language models' reasoning skills.

### F. THE PURPOSE OF THE RESEARCH

The primary aim of this work is to increase LLM performance on complex Ukrainian language exam tasks in a low-resource setup by employing parameter-efficient chain-of-thought fine-tuning. An important aspect of this research is to check whether, under resource constraints, enhanced fine-tuning and prompting methods can yield performance levels that rival those of larger proprietary models, ultimately advancing the application of cutting-edge information technologies in software engineering for educational domain.

This research includes the following tasks:

- development of a comprehensive baseline with parameter-efficient fine-tuning of selected open-source language models on a complete set of Ukrainian language exam problems, including multiple-choice and matching tasks;
- assessment of the impact of step-by-step reasoning by comparing models tuned solely for single-letter output with those tuned for chain-of-thought generation;
- comparison of the tuned models against leading proprietary and open-weight models.

## III. MATERIAL AND METHODS

### A. DATA PREPARATION

For training and evaluation, the complete Ukrainian language and literature dataset from the ZNO-Eval benchmark was used. This set consists of single-correct-answer questions and matching tasks, pairing numbered options with lettered options based on the question. The dataset combined 49 ZNO/EIE (External independent evaluation) and NMT exams, totaling 2,746 questions. The original ZNO-Eval task schema with the question, answer options, a correct answer, and a comment specifying the task topic was left unchanged (Fig. 1).

```
{
    "task_id": 8,
    "question": """Суфікс -ин- має однакове значення в усіх
    словах, ОКРІМ""",
    "answers": [
        {"answer": "А", "text": "соломина"},
        {"answer": "Б", "text": "бадилина"},
        {"answer": "В", "text": "височина"},
        {"answer": "Г", "text": "стеблина"}
    ],
    "answer_vheader": ["А", "Б", "В", "Г"],
    "answer_hheader": [],
    "correct_answer": ["В"],
    "comment": "ТЕМА: Словотвір. Суфіксальний спосіб.",
    "with_photo": False,
    "test_id": "522",
},
{
    "task_id": 27,
    "question": """З'ясуйте, якими частинами мови є виділені
    слова в реченні (цифра позначає наступне слово).\nСучасна
    людина, щоб бути (1)успішною, має вчитися (2)впродовж
    (3)усього життя, (4)опановуючи нові галузі знань.""",
    "answers": [
        {"answer": "А", "text": "займенник"},
        {"answer": "Б", "text": "прикметник"},
        {"answer": "В", "text": "форма дієслова (дієприкметник)"},
        {"answer": "Г", "text": "форма дієслова (дієприслівник)"},
        {"answer": "Д", "text": "прийменник"}
    ],
    "answer_vheader": ["А", "Б", "В", "Г", "Д"],
    "answer_hheader": ["1", "2", "3", "4"],
    "correct_answer": ["Б", "Д", "А", "Г"],
    "comment": "ТЕМА: Морфологія. Частини мови.",
    "with_photo": False,
    "test_id": "363"
}
```

Figure 1. ZNO-Eval schema for sample tasks from Table 1.

32 EIE tests were sampled for training, 13 EIE exams were chosen for validation, and 4 NMT exams were reserved for testing. The NMT exams were chosen for testing to align with the test set used in ZNO-Eval benchmarking and to avoid tasks requiring manual assessment. The training and validation sets included tasks from both the Ukrainian language and literature categories to evaluate generalization capabilities and prevent catastrophic forgetting caused by suboptimal tuning hyperparameters. The test set, however, contained only language tasks.

Prior to sampling, the dataset was cleaned by removing duplicate tasks (381), paraphrased tasks (52), tasks without answers (4), tasks missing a topic (48), and tasks containing photos in question or answer options (97). This preprocessing resulted in a final dataset of 1,740 tasks for training, 292 tasks for validation, and 108 tasks for testing.

### B. CHAIN-OF-THOUGHT DATA

This research proposes using both the task topic and a step-by-step solution as components for chain-of-thought tuning to enhance model performance and interpretability. This approach involves prefixing the generated solution with the corresponding task topic. The step-by-step solution aims to reduce hallucinations often observed in smaller models and improve interpretability by guiding the model to produce detailed reasoning rather than merely outputting a simple answer letter often seen in zero-shot prediction setups.

The inclusion of task topics mirrors the concept of generated knowledge prompting, where the model generates relevant contextual information to narrow the generation space. By prefixing solutions with task topics, the model is provided with additional context aimed at improving its ability to steer towards accurate and relevant outputs, keeping the model focused on the task-specific domain during result generation.

For each task, topics and step-by-step solutions were collected from the Osvita.ua portal [34], which contains educational materials and exam resources. Table 2 illustrates a sample task topic with its detailed step-by-step solution.

Table 2. Topic and solution for sample tasks from Table 1 with English translation

| Step-by-step solution | English translation of step-by-step solution |
|---|---|
| Коментар<br><br>ТЕМА: Словотвір. Суфіксальний спосіб.<br><br>Завдання перевіряє ваше вміння розпізнавати вивчені способи словотвору та аналізувати лексичне значення слова.<br><br>В українській мові за допомогою суфікса -ин- утворюють значну кількість іменників жіночого роду І відміни. Це слова на позначення частин рослини (бадилина, стеблина, соломина), а також на позначення території, рельєфу (височина).<br><br>**Відповідь – В.** | Comment<br><br>TOPIC: Word formation. Suffix-Based method.<br><br>This task tests your ability to recognize common word-formation processes and to analyze a word's lexical meaning.<br><br>In Ukrainian, the suffix -ин- is used to create many first-declension feminine nouns. These words either refer to plant parts (бадилина, стеблина, соломина) or to geographical features/terrain (височина).<br><br>**Answer – C.** |
| Коментар<br><br>ТЕМА: Морфологія. Частини мови.<br><br>Завдання перевіряє ваше вміння правильно визначати частини мови.<br><br>Необхідно бути дуже уважним, тому що частиномовна приналежність конкретного слова часто залежить від контексту.<br><br>До слова успішною можна поставити питання якою?, воно вказує на ознаку. Це **прикметник**.<br><br>До слова впродовж не можна поставити питання, воно лише служить для зв'язку слова життя з іншими в реченні. Це **прийменник**.<br><br>До слова усього можна поставити питання якого?, але воно лише вказує на ознаку, не називаючи її. Це **займенник**.<br><br>А слово опановуючи відповідає на питання що роблячи?, указує на додаткову дію. Це особлива форма дієслова **дієприслівник**.<br><br>**Відповідь – БДАГ.** | Comment<br><br>TOPIC: Morphology. Parts of speech.<br><br>This task tests your ability to correctly identify parts of speech.<br><br>It's important to be very attentive, because a word's part of speech often depends on the context.<br><br>You can ask "якою?" ("which one?") about "успішною" ("successful"), indicating a quality. That makes it an **adjective**.<br><br>You cannot form a question for "впродовж" ("throughout"); it simply links the word "життя" ("life") to other parts of the sentence. Therefore, it is a **preposition**.<br><br>You can ask "якого?" ("which one?") about "усього" ("all of"), but it only points to a characteristic without naming it. Hence, it is a **pronoun**.<br><br>The word "опановуючи" ("mastering" / "while mastering") answers "що роблячи?" ("while doing what?"), indicating an additional action. It is a special verb form called an **adverbial participle**.<br><br>**Answer – BEAD.** |

As shown in the figure above, after CoT tuning, the model is expected to generate a relevant hierarchical topic, prefixed with the keyword "ТЕМА:" ("TOPIC:"), followed by a detailed step-by-step solution. The solution includes a review of all answer options or pairs for the exam task, concluding with the keyword "Відповідь:" ("Answer:") and providing either a single answer letter for multiple-choice questions or a sequence of number-letter pairs for matching tasks. This structured approach ensures that the fine-tuned model delivers interpretable and accurate responses while maintaining alignment with task-specific requirements.

*C. DATA CONTAMINATION AND LEAKAGE*

Data contamination and leakage occur when information from the evaluation dataset inadvertently influences model training, leading to polluted performance metrics [35]. This problem questions the reliability of model evaluation, as it does not accurately reflect its ability to generalize to unseen data. Contamination, common for large language models trained on billions of texts, can arise from various sources, such as shared content between datasets or pre-training on datasets containing evaluation tasks.

In this research, two types of data contamination and leakage were addressed. Pre-training data contamination explores the possibility that the large language model was pre-trained on test-exam tasks. However, this issue is mitigated by several factors. The availability of webpages with Ukrainian exam data is limited, and Ukrainian was not a primary language in the LLM's pre-training dataset. Furthermore, in most cases, the correct answer or problem solution is not directly available alongside the question definition. Accessing the solution often requires additional actions, such as logging in or revealing answers embedded as images rather than text.

To further reduce the impact of potential contamination on evaluation results, the answer numbers, letters, and texts were shuffled for the test set. This measure prevents straightforward answer memorization from contaminating the results.

The second type aims to check whether some tasks within the dataset contained exact or partial matches of questions or answer options across training, validation, and test sets. These duplicates were identified based on matching questions or answer option text in case they were not common generic statements. Generic statements, such as "match options on the left with texts on the right" for questions or part-of-speech keywords for answers, were excluded from duplicate identification.

All duplicates between the validation and training sets were removed from the validation set. Similarly, duplicates between the test set and either the training or validation sets were removed from the train/validation to ensure that the test set remained unchanged. Final manual sample screening identified 40 tasks in the validation set and 12 tasks in the test set that contained rephrased questions or answer options. Although these instances are not exact duplicates, they were removed to prevent data leakage and minimize any potential impact on the evaluation scores.

### D. MODEL SELECTION

To align with the low-resource goal of this research, the selection of models was limited to compact options that could be efficiently trained on a single A100 GPU with 80G of VRAM. The chosen models include Meta's LLaMA 3.1 with 8 billion parameters, LLaMA 3.2 with 3 billion parameters, and Gemma 2 with 9 billion parameters. These models were selected due to their balance between performance and computational efficiency, making them suitable for resource-constrained environments.

Each model comes in two versions: base pre-trained and fine-tuned with instructions to follow user commands in a chat-like manner. This research focuses solely on instruction-tuned versions since pre-trained models did not provide any improvement during initial evaluations.

To further optimize training and inference processes, all models were quantized to 4-bit with the Bits and Bytes library [36]. This quantization significantly reduced memory usage, enabling faster training and inference. The combination of compact architecture and quantization ensured efficient use of computational resources, allowing for effective experimentation under low-resource conditions.

### E. MODEL TUNING AND EVALUATION

Parameter-efficient fine-tuning was performed on the selected instruction-tuned models ("it" in the model name) versions, using two variations: one with the correct answer represented as a letter or sequence of letters and the other incorporating the proposed chain-of-thought approach with and without topics. Models were fine-tuned over four epochs, with a learning rate of 3e-04 and checkpoints saved after each epoch. The best checkpoint was identified based on the validation metric that produced the highest overall score on validation exams.

The loss was not used as the validation metric because it is based on the model's perplexity, which does not account for the importance of generating the correct answer letter. Perplexity treats all characters in the generation equally and does not consider the variability in phrasing step-by-step solutions. Instead, validation accuracy, calculated as the sum of all scores on the validation exams, was used to find the best checkpoint.

The gradient accumulation technique helped mimic large batch sizes on GPUs with limited memory during fine-tuning. Also, the tuning process utilized 4-bit precision models along with LoRA of rank 16. Data preparation, model configs, and PEFT scripts are available at *github.com/NLPForUA/ZNO*.

Table 3 shows all promising experimental parameters.

**Table 3. Experiment parameters**

| Model | Tuning | Parameters, billions | Trained parameters, millions | Batch size | Accumulation |
|---|---|---|---|---|---|
| Tuned for answer letter generation | | | | | |
| LLaMA-3.2-3B-it-tune-al | letter | 3 | 22 | 8 | 4 |
| LLaMA-3.1-8B-it-tune-al | letter | 8 | 44 | 4 | 4 |
| Gemma-2-9B-it-tune-al | letter | 9 | 52 | 4 | 4 |
| Tuned for chain-of-thought (step-by-step solution) generation | | | | | |
| LLaMA-3.2-3B-it-tune-cot | solution | 3 | 22 | 8 | 4 |
| LLaMA-3.1-8B-it-tune-cot | solution | 8 | 44 | 4 | 4 |
| Gemma-2-9B-it-tune-cot | solution | 9 | 52 | 4 | 4 |
| Tuned for chain-of-thought (topic and step-by-step solution) generation | | | | | |
| LLaMA-3.2-3B-it-tune-cot-wt | topic + solution | 3 | 22 | 8 | 4 |
| LLaMA-3.1-8B-it-tune-cot-wt | topic + solution | 8 | 44 | 4 | 4 |
| Gemma-2-9B-it-tune-cot-wt | topic + solution | 9 | 52 | 4 | 4 |

For evaluation, baseline scores were established using random guessing and zero-shot evaluations of models without CoT output. The evaluation used greedy decoding with a maximum generation length of 2,048 tokens. Generated answers were extracted from the last occurrence of the "Відповідь:" ("Answer:") keyword. The scoring approach followed the same rules for both EIE and NMT exams. Multiple-choice questions were scored 1 point for each correct prediction, while matching tasks were scored up to 4 points, with 1 point awarded for each correct logical pair. For single-answer questions, a score of zero was given if multiple letters were generated, even if the first answer was correct. The score for the matching task was also zeroed if the response contained more than four answer letters, motivating confident solution generation. This methodology ensured consistent evaluation across all models and tasks.

## IV. RESULTS

In general, the experimental results prove the effectiveness of parameter-efficient fine-tuning combined quantization for compact open-source models. For all configurations, tuned models demonstrated substantial improvements over the baseline, with joint topic generation and step-by-step reasoning contributing moderately to performance gains. Detailed experiment results are demonstrated in Tables 4 and 5. All tuned models are available at *huggingface.co/NLPForUA*.

**Table 4. Evaluation results on validation set**

| Model Name | Generates | Scores for language tests | | | Scores for language and literature | | |
|---|---|---|---|---|---|---|---|
| | | Single answer | Matching | Total | Single answer | Matching | Total |
| Max possible score | - | 233 | 72 | 305 | 260 (+27) | 88 (+16) | 348 |
| Random guess | letter | 53.3 | 14.4 | 67.7 | 58.8 (+5.5) | 17.6 (+3.2) | 76.4 |
| Baseline: zero-shot answer letter generation | | | | | | | |
| LLaMA-3.2-3B-it | letter | 0 | 1 | 1 | 1 (+1) | 1 (+0) | 2 |
| Qwen2.5-7B-it | letter | 52 | 5 | 57 | 57 (+5) | 8 (+3) | 65 |
| LLaMA-3.1-8B-it | letter | 66 | 10 | 76 | 71 (+5) | 11 (+1) | 82 |
| Gemma-2-9B-it | letter | 31 | 16 | 47 | 36 (+5) | 18 (+2) | 54 |
| Qwen2.5-14B-it | letter | 69 | 16 | 85 | 81 (+12) | 19 (+3) | 100 |
| Gemma-2-27B-it | letter | **79** | 20 | 99 | **88** (+9) | **22** (+2) | **110** |
| Qwen2.5-32B-it | letter | 40 | 12 | 52 | 48 (+8) | 16 (+4) | 64 |
| LLaMA-3.3-70B-it | letter | 56 | 15 | 71 | 64 (+8) | 18 (+3) | 82 |
| Qwen2.5-72B-it | letter | 61 | 12 | 73 | 74 (+13) | 14 (+2) | 88 |
| Reasoning models baseline: zero-shot chain-of-thought generation | | | | | | | |
| DeepSeek-R1 LLaMA-8B | solution | 9 | 0 | 9 | 11 (+2) | 0 (+0) | 11 |
| DeepSeek-R1 Qwen-14B | solution | 25 | 13 | 38 | 35 (+10) | 13 (+0) | 48 |
| DeepSeek-R1 Qwen-32B | solution | 43 | **29** | 72 | **51** (+8) | **29** (+0) | **80** |
| LLaMA 3.2 3B | | | | | | | |
| LLaMA-3.2-3B-it-tune-al | letter | 57 | 16 | 73 | **65** (+8) | 17 (+1) | **82** |
| LLaMA-3.2-3B-it-tune-cot | solution | 54 | 17 | 71 | 63 (+9) | **18** (+1) | 81 |
| LLaMA-3.2-3B-it-tune-cot-wt | topic+solution | 53 | 8 | 61 | 60 (+7) | 13 (+5) | 73 |
| LLaMA 3.1 8B | | | | | | | |
| LLaMA-3.1-3B-it-tune-al | letter | 74 | 27 | 101 | 82 (+8) | 31 (+4) | 113 |
| LLaMA-3.1-8B-it-tune-cot | solution | 82 | 28 | 110 | **94** (+12) | 32 (+4) | 126 |
| LLaMA-3.1-8B-it-tune-cot-wt | topic+solution | 81 | 35 | 116 | 91 (+10) | **38** (+3) | **129** |
| Gemma 2 9B | | | | | | | |
| Gemma-2-9B-it-tune-al | letter | 104 | 37 | 141 | **118** (+14) | 41 (+4) | **159** |
| Gemma-2-9B-it-tune-cot | solution | 96 | 41 | 137 | 109 (+13) | **44** (+3) | 153 |
| Gemma-2-9B-it-tune-cot-wt | topic+solution | 94 | 37 | 131 | 110 (+16) | 39 (+2) | 149 |

The added benefit of chain-of-thought tuning (LLaMA and Gemma models with "cot" suffix) becomes clearer when applied to more complex tasks, including matching and literature assignments (scores shown in parentheses for literature tasks in Table 4). In these scenarios, the implementation of step-by-step reasoning enhances the steerability and clarity of the model's thought process, making it easier to follow the logic it employs to arrive at conclusions. However, despite these gains, the validation set did not consistently show anticipated improvement when comparing chain-of-thought to letter-only generation. Several factors appear to affect the result. Firstly, the validation set primarily consists of older exam tasks with no answer option shuffling, thus increasing the chance of data contamination. Secondly, the approach taken to remove duplicate and rephrased tasks has inadvertently led to an uneven distribution of task types and topics. Some appear only once or twice, whereas others are overrepresented. Lastly, adapters were merged with base models for validation scoring due to time and cost considerations. This could lead to a score drop for CoT models.

Nevertheless, the validation scores remain valuable for selecting the optimal training epoch. It has been empirically observed that higher single-answer, matching, and total validation scores directly correlate with better performance on a more representative test set. In contrast to the validation set, the test data includes more recent exams with answer options shuffling and a fair balance of question types and topics.

Table 5. Evaluation results on test set

| Model Name | Generates | Total scores | | | Total scores after merge | | |
|---|---|---|---|---|---|---|---|
| | | Single answer | Matching | Total | Single answer | Matching | Total |
| Max possible score | - | 92 | 64 | 156 | - | - | - |
| Random guess | letter | 20.25 | 12.78 | 33.03 | - | - | - |
| Baseline: zero-shot answer letter generation | | | | | | | |
| LLaMA-3.2-3B-it | letter | 0 | 4 | 4 | - | - | - |
| Qwen2.5-7B-it | letter | 26 | 5 | 31 | - | - | - |
| LLaMA-3.1-8B-it | letter | 25 | 7 | 32 | - | - | - |
| Gemma-2-9B-it | letter | 21 | 21 | 42 | - | - | - |
| Qwen2.5-14B-it | letter | 25 | 16 | 41 | - | - | - |
| Gemma-2-27B-it | letter | **30** | 24 | **54** | - | - | - |
| Qwen2.5-32B-it | letter | 18 | **26** | 44 | - | - | - |
| LLaMA-3.3-70B-it | letter | 25 | 13 | 38 | - | - | - |
| Qwen2.5-72B-it | letter | 18 | 15 | 33 | - | - | - |
| Reasoning models baseline: zero-shot chain-of-thought generation | | | | | | | |
| DeepSeek-R1 LLaMA-8B | solution | 4 | 1 | 5 | - | - | - |
| DeepSeek-R1 Qwen-14B | solution | 16 | 21 | 37 | - | - | - |
| DeepSeek-R1 Qwen-32B | solution | **22** | **25** | **47** | - | - | - |
| LLaMA 3.2 3B | | | | | | | |
| LLaMA-3.2-3B-tune-al | letter | 24 | 11 | 35 | **27** | 10 | 37 |
| LLaMA-3.2-3B-it-tune-cot | letter | 18 | 10 | 28 | 16 | 14 | 30 |
| LLaMA-3.2-3B-it-tune-cot-wt | topic+solution | 24 | **15** | **39** | 14 | 5 | 19 |
| LLaMA 3.1 8B | | | | | | | |
| LLaMA-3.1-3B-it-tune-al | letter | 25 | 12 | 37 | **30** | **17** | **47** |
| LLaMA-3.1-8B-it-tune-cot | solution | 19 | 13 | 32 | 26 | 13 | 39 |
| LLaMA-3.1-8B-it-tune-cot-wt | topic+solution | 26 | 15 | 41 | 28 | 14 | 42 |
| Gemma 2 9B | | | | | | | |
| Gemma-2-9B-it-tune-al | letter | **33** | 23 | 56 | 41 | 22 | 63 |
| Gemma-2-9B-it-tune-cot | solution | 37 | 27 | **64** | 28 | **29** | 57 |
| Gemma-2-9B-it-tune-cot-wt | topic+solution | 29 | **30** | 59 | 28 | 26 | 54 |

The random guessing baseline, selecting one random answer out of all provided options for questions with a single correct answer (multiple-choice question) and constructing a sequence of four random letters for matching tasks, achieved a total test score of 33.03, with 20.25 points on single-answer questions and 12.78 points on matching tasks. The overall performance of baseline LLaMA-3.2-3B-it and LLaMA-3.1-8B-it reflects the underrepresented nature of the Ukrainian language in pre-training datasets, as the former fails to provide any meaningful answer (total score of 4) and the latter struggles to surpass random guessing in matching tasks (7 vs 12.78 points). At the same time, base Gemma-2-9B-it demonstrated high robustness without any fine-tuning, securing 21 points on average for matching tasks and 42 in total. In comparison, its 3 times larger "relative" became a leader with 54 points.

An important consideration is the effect of merging 4-bit LoRA adapter weights with the base model. Directly merging in 4-bit often degrades prediction quality due to rounding errors and precision loss. Another approach, a full precision merge with a subsequent quantization, helped mitigate the issue. Interestingly, letter-only models show substantial gains after merging as the impact of numerical artifacts increases with the length of the generated sequence.

All instruct models tuned to generate answer letters or sequences of letters (models with the "it-tune-al" suffix) demonstrated reasonable improvements over the baseline. For instance, the LLaMA-3.2-3B model tuned with topics and solutions slightly exceeded the random guessing score (39 vs. 33 points), while LLaMA-3.1-8B-it-tune-al yielded 49.9% improvement with a total score of 47 points, and Gemma-2-9B-tune-al demonstrated the second best result across Gemma models with 63 points.

Fine-tuning models with step-by-step reasoning solutions in a chain-of-thought manner revealed nuanced outcomes. While the CoT approach substantially improved performance across all matching and literature tasks compared to letter-only solutions, only Gemma-2-9B-it-tune-cot achieved a higher total score with 14.3% and 1.6% gains over the letter-only solution before and after model merge.

A proposed method, prefacing step-by-step reasoning with a generated task topic, helped squeeze the solution space, enhancing both interpretability and accuracy. With this approach, the topic is kind of a knowledge prompt, guiding the model to focus on relevant aspects of the task. That pseudo-knowledge injection increases the likelihood of the domain-specific terms and rules, helping the model "recall" information memorized during pre-training or instruction tuning. Joint topic and step-by-step solution generation ("cot-wt" suffix) allowed to surpass the 30-point bar for matching tasks and secure 18 points for literature tasks with Gemma model, while also demonstrating the highest score of 39 points across all LLaMA-3.2-3B solutions with 5.4% and 30% increase over LLaMA-3.2-3B-tune-al and LLaMA-3.2-3B-it-tune-cot.

Although chain-of-thought enhancements over a letter-only generation are not always dramatic, this approach has unquestionable benefits with improved matching scores and interpretability (Table 6).

Table 6. Solutions for sample tasks by Gemma-2-9B-it-tune-cot-wt and their translations

| Generated chain-of-thought solution | English translation of generated chain-of-thought solution |
|---|---|
| <bos><start_of_turn>user<br>Дайте розгорнуту відповідь на завдання, починаючи з ключового слова "Відповідь:" та використовуючи лише наведені нижче варіанти.<br><br>Завдання: Суфікс -ин- має однакове значення в усіх словах, ОКРІМ<br><br>Варіанти відповіді:<br>А - соломина<br>Б - бадилина<br>В - височина<br>Г - стеблина<end_of_turn><br><start_of_turn>model<br>Відповідь:<br>ТЕМА: Будова слова. Словотвір. Значущі частини слова: корінь, префікс, суфікс, закінчення<br>Для виконання цього завдання треба пригадати поняття значущих частин слова (морфем), зокрема кореня й суфікса. Суфікс -ин- має різне значення. Так у словах соломина, стеблина, бадилина він позначає відірване від чогось, відбиток, уламки, дрібні частинки чогось.<br>А от у слові височина цей суфікс має значення збірності.<br>Відповідь – В.<end_of_turn> | <bos><start_of_turn>user<br>Provide a detailed answer to the task, starting with the keyword "Відповідь:" and using only the options listed below.<br><br>Task: The suffix -ин- has the same meaning in all of the following words EXCEPT<br><br>Answer options:<br>A – соломина (straw piece)<br>B – бадилина (leafy-stalk piece)<br>C – височина (height / highland)<br>D – стеблина (stem piece) <end_of_turn><br><start_of_turn>model<br>TOPIC: Word Structure. Word Formation. Meaningful Parts of the Word: Root, Prefix, Suffix, Ending<br>To complete this task, you need to recall the concept of the meaningful parts of a word (morphemes), particularly the root and the suffix. The suffix -ин- can have different meanings. For example, in the words "соломина", "стеблина", and "бадилина", it indicates something torn off, a fragment, remainder, or small part of something.<br>However, in the word "височина", this suffix signifies collectivity.<br>Answer – B. <end_of_turn> |
| <bos><start_of_turn>user<br>Дайте розгорнуту відповідь на завдання, починаючи з ключового слова "Відповідь:" та використовуючи лише наведені нижче варіанти.<br><br>Завдання: З'ясуйте, якими частинами мови є виділені слова в реченні (цифра позначає наступне слово).<br>Сучасна людина, щоб бути (1)успішною, має вчитися (2)впродовж (3)усього життя, (4)опановуючи нові галузі знань.<br><br>Варіанти відповіді:<br>А – займенник<br>Б – прикметник<br>В – форма дієслова (дієприкметник)<br>Г – форма дієслова (дієприслівник)<br>Д – прийменник<end_of_turn><br><start_of_turn>model<br>Відповідь:<br>ТЕМА: Морфологія. Частини мови<br>Завдання перевіряє ваше вміння визначати частини мови. Необхідно бути дуже уважним, тому що частиномовна приналежність конкретного слова часто залежить від контексту. Проаналізуймо кожне із запропонованих слів.<br>Слово **успішною** відповідає на питання яка? та вказує на ознаку предмета. Отже, це прикметник.<br>Слово **впродовж** не відповідає на питання, а лише поєднує слова в реченні. Отже, це прийменник.<br>Слово **усього** відповідає на питання що? і вказує на предмет, але не називає його. Отже, це займенник.<br>А от слово **опановуючи** відповідає на питання що роблячи? та вказує на додаткову дію підмета. Отже, це особлива форма дієслова дієприслівник.<br>Відповідь: 1 – Б, 2 – Д, 3 – А, 4 – Г.<end_of_turn> | <bos><start_of_turn>user<br>Provide a detailed answer to the task, starting with the keyword "Відповідь:" and using only the options listed below.<br><br>Task: Determine which parts of speech the highlighted words are in the following sentence (the number indicates the word that follows).<br>A modern person, in order to be (1)successful, must keep studying (2)throughout (3)their entire life, (4)mastering new fields of knowledge.<br><br>Answer options:<br>A – pronoun<br>B – adjective<br>C – verb form (participle)<br>D – verb form (adverbial participle)<br>E – preposition<end_of_turn><br><start_of_turn>model<br>Answer:<br>TOPIC: Morphology. Parts of speech<br>This task checks your ability to identify parts of speech. You must be very attentive, because the part of speech for a given word often depends on the context. Let's analyze each of the proposed words.<br>The word **successful** answers the question "which one?" and indicates a characteristic of an object. So, it is an adjective.<br>The word **throughout** does not answer a question; it merely connects words in the sentence. Therefore, it is a preposition.<br>The word **their** answers the question "whose?" and indicates an object without naming it. So, it is a pronoun.<br>The word **mastering** answers the question "doing what?" and indicates an additional action of the subject. Therefore, it is a special verb form known as an adverbial participle.<br>Answer: 1 – B, 2 – E, 3 – A, 4 – D.<end_of_turn> |

Gemma's solution presented above demonstrates several strengths but also has some limitations. In the first task, the model seems to apply a deep linguistic analysis with strong reasoning. However, in the second task, the generated answer explains the reasoning behind each matching decision, clarifying the pairing of specific fragments without going deeper into the morphological aspects behind each answer option. Both answers are correct, so this result is still substantial as less than 50% of graduates select a correct option in tasks like the first problem, and only 28%, on average, strike four out of four in problems similar to the second one [37].

In addition to a comparison between open-weight LLMs, it is also crucial to check how close the obtained solutions are to leading proprietary models widely used by the community and enterprise. Fig. 2 presents a combined result chart of tuned models and zero-shot LLMs.

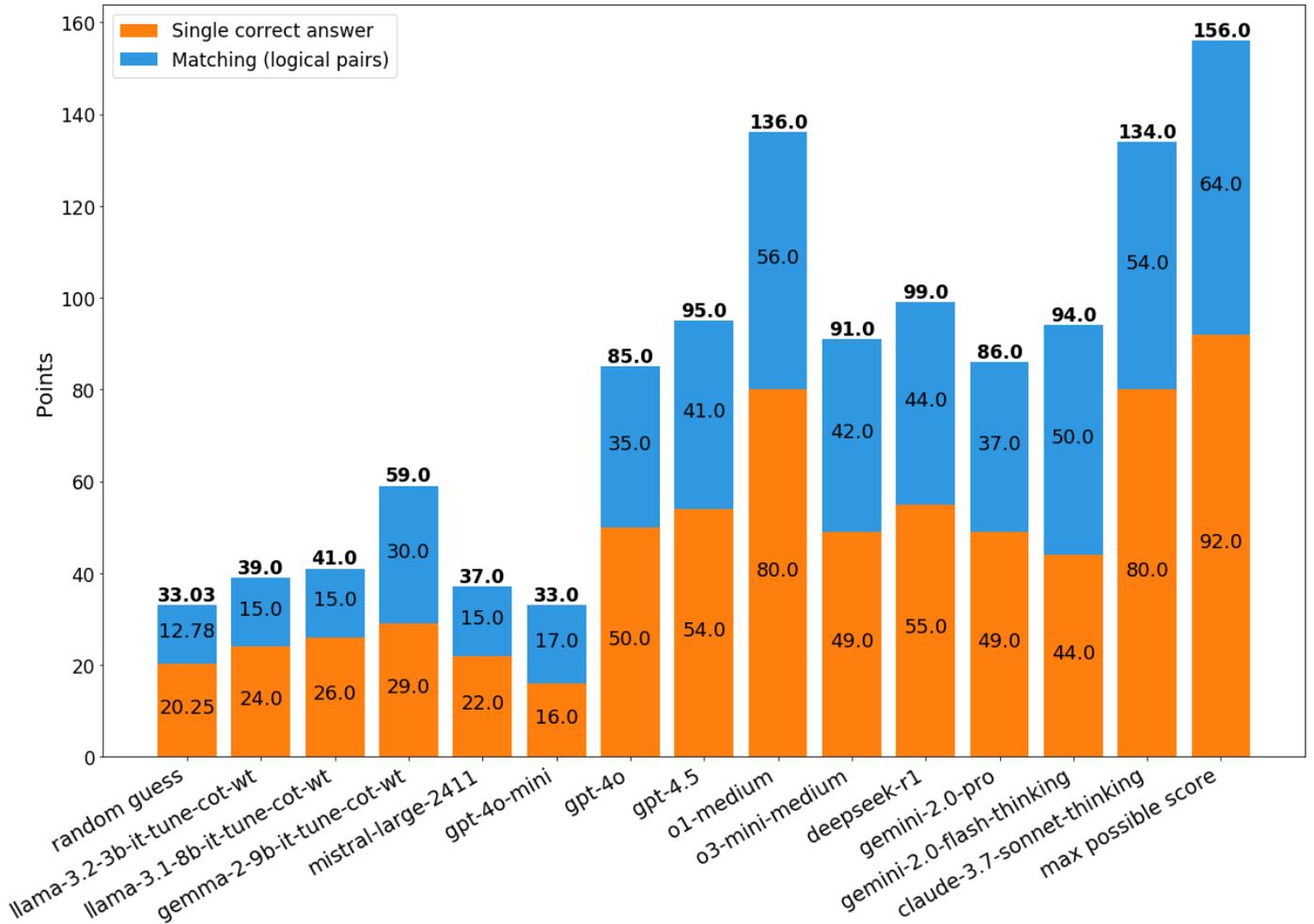

Figure 2. Evaluation results for tuned open-weight and zero-shot proprietary models

Despite these advancements, compact models still cannot reach the performance level of leading reasoning models like OpenAI o1 or Claude 3.7 Sonnet, which benefit from extensive multilingual datasets and demonstrate strong results in complex tasks. However, fine-tuned models highlight that combining parameter-efficient fine-tuning with CoT reasoning could significantly narrow the performance gap in a low-resource setup and even slightly outperform larger LLMs (GPT-4o mini and Mistral Large). Moreover, Gemma secured 30 points for matching tasks, getting relatively close to powerful GPT-4o and Gemini 2.0 Pro models (35 and 37 points).

## V. CONCLUSIONS

This research provides several important contributions to the field of natural language processing, particularly for low-resource setups and underrepresented languages. Furthermore, to the best of our knowledge, this work represents the first comprehensive evaluation of large language models on matching tasks for Ukrainian language exams and extends the Ukrainian language exam benchmark with common open-weight and proprietary reasoning models.

The scientific novelty of the obtained result lies in the proposed method of joint parameter-efficient fine-tuning and step-by-step reasoning with task-specific knowledge generation. This method not only provides substantial quality improvement, interpretability, and robustness compared to standard answer letter generation and chain-of-thought tuning in reasoning-intensive Ukrainian exam tasks for open-weight LLaMA and Gemma models but also underscores the potential for cost-effective alternatives to proprietary LLMs.

The practical significance of the research is the demonstration of how compact models can be optimized to perform well on complex tasks in low-resource environments. By using a single A100 GPU, LoRA, and 4-bit quantization techniques, the work underscores the possibility of training advanced NLP systems in computationally constrained environments. The findings are particularly relevant for underrepresented languages, where access to proprietary models and computational resources may be limited.

The limitation of this research is that the evaluation data size, though representative, is relatively small and may not fully capture the diversity of real-world tasks. Additionally, unavoidable data contamination during pre-training and the limited hyperparameter exploration in the experiments could influence the generalization of the obtained results. Moreover, the use of 4-bit quantization, while beneficial for efficiency, might also introduce subtle degradation in model performance, which requires further exploration.

The prospect for further research is to mitigate the aforementioned limitations by expanding the evaluation dataset to include more diverse tasks, exploring multimodal reasoning capabilities, and experimenting with a broader range of hyperparameters.

## VI. ACKNOWLEDGEMENTS